\documentclass{article}

\usepackage[dblblindworkshop,final]{neurips_2026}
\usepackage{amsmath}
\usepackage{amssymb}
\usepackage{booktabs}
\usepackage[T1]{fontenc}
\usepackage{graphicx}
\usepackage{microtype}
\usepackage{tabularx}
\usepackage{url}
\usepackage{xcolor}
\usepackage{hyperref}

\hypersetup{
  colorlinks,
  linkcolor={red!50!black},
  citecolor={blue!50!black},
  urlcolor={blue!80!black}
}

\workshoptitle{Long-Context Foundation Models}

\title{CoRA-Bench: Evaluating Evidence Preservation by Context Selectors Under Token Budgets}

\author{%
Huzaifa Pardawala\textsuperscript{1}
\quad
Siddhant Sukhani\textsuperscript{1}
\quad
Sudheer Chava\textsuperscript{2}
\quad
Agam Shah\textsuperscript{3}
\\[7pt]
{\small
\textsuperscript{1}Stanford University
\qquad
\textsuperscript{2}Georgia Institute of Technology
\qquad
\textsuperscript{3}ZettaQuant
}
}

\begin{document}
\maketitle

\begin{abstract}
Context selection can reduce input length during long-input inference but can remove evidence required to answer a query. We introduce \textsc{CoRA-Bench}\footnote{\label{fn:code}Code repository: \url{https://github.com/ZettaQuant/cora-bench}.}, a benchmark for evaluating evidence preservation under shared token budgets. We compare head and tail truncation, BM25, dense retrieval, dense retrieval with reranking, LLMLingua-2, and Provence across BABILong, LOONG Financial Spotlight, a custom NoLiMa diagnostic, and LongMemEval. Evidence-preservation results reveal task-dependent tradeoffs: BM25 achieves 0.921 supporting-fact recall on BABILong at 128K, dense retrieval and dense reranking each achieve 0.908 needle recall on NoLiMa at 64K, and dense retrieval achieves 0.993 answer-bearing-turn recall on LongMemEval at 64K. Evidence, token retention, and complementary reader evaluations use the same selected contexts. Reader evaluations use dataset-native answer metrics and three runs on the same questions, with paired bootstrap confidence intervals. At a 32K budget, BM25 achieves 46.20\% BABILong answer accuracy and dense retrieval achieves 71.13\% LongMemEval accuracy. The results show that method rankings vary across evidence types and between evidence recall and downstream accuracy.
\end{abstract}

\section{Introduction}

Performance on long-context tasks depends on evidence position, even within supported context windows \citep{liu2024lost}. Processing full reports or conversation histories also increases input length when much of the content may be irrelevant. Context selection reduces this input but risks discarding supporting evidence.

We compare selectors under shared token budgets, measuring evidence recall, token retention, and downstream answer accuracy on the same reader-visible contexts.

Our contribution is a common implementation and evaluation protocol for six practical selectors across four complementary long context settings. BABILong provides explicit supporting facts in long book text \citep{kuratov2024babilong}. LOONG contributes evidence localization in multi-document financial contexts \citep{wang2024loong}. NoLiMa tests retrieval with minimal literal overlap \citep{modarressi2025nolima}. LongMemEval supplies answer-bearing turns in multi-session dialogue \citep{wu2025longmemeval}. Across these settings, CoRA-Bench provides common token accounting and budget enforcement, occurrence-aware evidence tracing, and direct comparison at 32K, 64K, and, where source length permits, 128K tokens.

We evaluate downstream answers using Gemini 2.5 Flash-Lite with dataset-native scoring and paired bootstrap confidence intervals (Appendix~\ref{app:reader-matrix}).

\section{Related work}

Long context benchmarks cover position sensitivity, broad task collections, controlled synthetic tests, multi-document reasoning, latent retrieval, and conversational memory \citep{liu2024lost,bai2023longbench,bai2024longbenchv2,hsieh2024ruler,kuratov2024babilong,li2024needlebench,wang2024loong,modarressi2025nolima,maharana2024locomo,wu2025longmemeval, tatarinov-etal-2026-kg}. This diversity motivates evaluating selectors across evidence regimes instead of treating long context retrieval as one task.

Context reduction spans lexical search and reproducible retrieval toolkits \citep{robertson2009probabilistic,lin2021pyserini}, dense passage retrieval \citep{karpukhin2020dense}, retrieval augmented generation and neural reranking \citep{lewis2020rag,nogueira2019passage,yu2024rankrag}, and prompt or retrieved-context compression \citep{jiang2023llmlingua,jiang2024longllmlingua,pan2024llmlingua2,xu2024recomp,chirkova2025provence}. CoRA-Bench contributes the controlled evaluation layer needed to compare these heterogeneous approaches under common token budgets and evidence definitions.

Learned representation compression includes gist tokens \citep{mu2024learningcompresspromptsgist}, AutoCompressors \citep{chevalier2023adaptinglanguagemodelscompress}, and the In-context Autoencoder \citep{ge2024incontextautoencodercontextcompression}. Perception Compressor combines question-guided retrieval with token pruning \citep{tang2025perceptioncompressortrainingfreeprompt}; COMI allocates compression across groups and merges tokens using marginal information gain \citep{tang2026comicoarsetofinecontextcompression}. These approaches broaden the compression landscape. Our evaluation focuses on selectors that emit text for a shared reader.

Architectural and systems advances have improved the feasibility of long-context inference \citep{vaswani2017attention,dao2024flashattention2,kwon2023pagedattention,geminiteam2024gemini15}. We examine evidence preservation under token budgets using a fixed downstream reader.

\begin{figure*}[t]
\centering
\includegraphics[width=\textwidth]{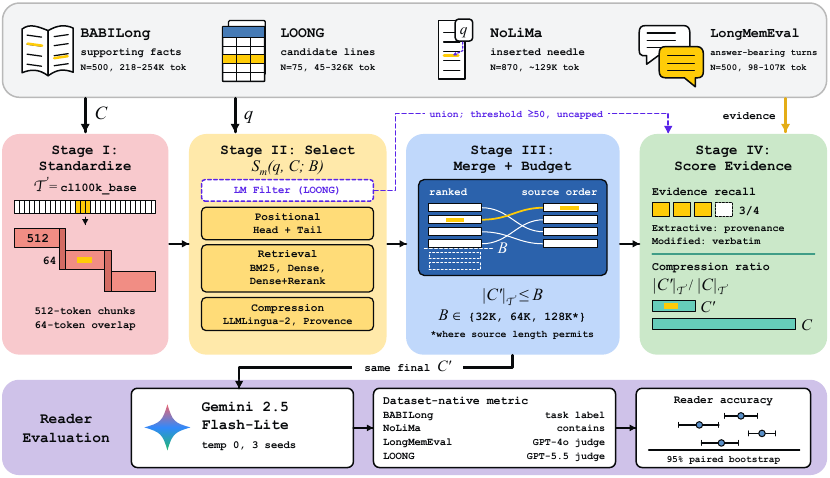}
\caption{CoRA-Bench: shared token budgets and dataset-specific evidence scoring across four retrieval regimes. Evidence and reader evaluation use the same final context, with overlapping retrieval spans merged; reader inference is repeated with 3 seeds. The LM-filter bypass denotes Threshold $\geq50$, uncapped; its ranked, capped arms follow budget enforcement without a score threshold.}
\label{fig:methodology}
\end{figure*}

\section{Evaluation}
\label{sec:evaluation}

Figure~\ref{fig:methodology} summarizes the protocol. Given a query $q$, source context $C$, budget $B$, and selector $S_m$, we construct
\begin{equation}
  C' = S_m(q,C;B), \qquad |C'|_{\mathcal{T}} \leq B,
\end{equation}

Here, $\mathcal{T}$ denotes the \texttt{cl100k\_base} tokenizer from \texttt{tiktoken} \citep{openai2022tiktoken}. We use it for chunking and token accounting, although it may differ from model-native tokenizers. The 32K, 64K, and 128K budgets (K = 1,000) cap selected context, excluding query and instruction overhead. We report macro-averaged compression ratios $|C'|_{\mathcal{T}}/|C|_{\mathcal{T}}$ and question-level evidence recall over evidence-evaluable examples. Extractive methods use source-occurrence provenance and text-modifying methods require verbatim evidence survival. Full context has a recall of one by construction.

\paragraph{Budgets and cohorts.}
BABILong and NoLiMa use the same examples at every reported budget. LongMemEval sources are shorter than 128K, so only 32K and 64K are informative. For LOONG, we follow the existing run rule that evaluates only sources longer than the budget, giving 75, 70, and 51 examples for reader accuracy and token ratios at 32K, 64K, and 128K. Evidence-evaluable counts are 74, 69, and 50; the missing candidate annotation is excluded from recall, not treated as a selection failure. LOONG values should therefore be read within a budget; changes across budgets also reflect cohort changes.

\begin{table*}[t]
\centering
\caption{Evaluated data. Lengths use \texttt{cl100k\_base}. NoLiMa is a custom repeated-measures diagnostic, not the official benchmark aggregate. LongMemEval is treated as static context selection, not online memory management.}
\label{tab:datasets}
\small
\setlength{\tabcolsep}{3.5pt}
\renewcommand{\arraystretch}{1.08}
\begin{tabularx}{\textwidth}{@{}l>{\raggedright\arraybackslash}p{2.5cm}c>{\raggedright\arraybackslash}p{2.5cm}X@{}}
\toprule
\textbf{Dataset} & \textbf{Selection stress} & \textbf{N} & \textbf{Source tokens} & \textbf{Configuration} \\
\midrule
BABILong \citep{kuratov2024babilong} & Sparse supporting facts & 500 & 217.7K to 253.7K & qa1 to qa5, 100 examples each, generated at 256K with gold support occurrences \\
LOONG \citep{wang2024loong} & Financial line localization & 75 & 45.2K to 326.4K & English Financial Spotlight, level 1; 2 to 9 documents; 74/69/50 evidence-evaluable at 32K/64K/128K \\
NoLiMa \citep{modarressi2025nolima} & Minimal lexical overlap & 870 & 128.57K to 128.74K & 58 question and needle pairs, 3 depths, 5 fixed haystacks; repeated measures \\
LongMemEval \citep{wu2025longmemeval} & Conversational retrieval & 500 & 98.4K to 107.1K & Cleaned small split; 470 answerable and 30 abstention examples \\
\bottomrule
\end{tabularx}
\end{table*}

\paragraph{Methods.}
All retrievers use 512 token chunks with 64 token overlap and return selected chunks in source order. Head and tail truncation is a query independent positional baseline. BM25 uses Okapi scoring \citep{robertson2009probabilistic}. Dense retrieval uses Qwen3-Embedding-0.6B, and Dense plus Rerank applies Qwen3-Reranker-0.6B to candidate pools of up to 100, 133, and 258 chunks at 32K, 64K, and 128K \citep{zhang2025qwen3embedding}. LLMLingua-2 performs query-agnostic token compression \citep{pan2024llmlingua2}. Provence reranks and prunes sentences within chunks at threshold 0.1, without forced title retention \citep{chirkova2025provence}. The neural selectors use PyTorch and the Transformers ecosystem \citep{paszke2019pytorch,wolf2020transformers}. BM25, dense retrieval, and dense reranking merge overlapping source-token spans, accept a strict prefix of ranked chunks under the output budget, and restore source order. Reranking can underfill the budget when its candidate pool is exhausted. Evidence, token, and reader scores use these same final outputs. Final emitted text is re-counted under $\mathcal{T}$ and must satisfy $B$. Text-modifying methods are evaluated on their actual emitted text. Financial tables retain their materialized source representation, without table-specific preservation or whitespace normalization.

\paragraph{Dataset specific evidence.}
BABILong uses occurrence-aware recall of gold supporting facts. LOONG uses automatic candidate-line survival, unless independently reviewed support spans are supplied. These candidate lines are not human-annotated gold evidence. This diagnostic is distinct from LOONG's published document coverage Recall@$n$ and from its judge based answer score. NoLiMa uses verbatim survival of the inserted needle. LongMemEval uses recall of annotated answer bearing turns; its 30 abstention questions have no positive evidence and are excluded from evidence recall.

\section{Results}

\paragraph{Evidence preservation.}
Rankings depend on the retrieval regime. BM25 leads BABILong supporting-fact recall at all three budgets (0.838, 0.868, 0.921). At 64K, dense retrieval and dense reranking each achieve 0.908 NoLiMa needle recall versus 0.467 for BM25. Dense reranking retains 0.979 and 0.991 of LongMemEval answer-bearing turns at 32K and 64K; dense retrieval reaches 0.993 at 64K. These findings favor lexical retrieval under explicit surface cues and semantic retrieval under limited lexical overlap (Appendix~\ref{app:results}).


\begin{table}[t]
\centering
\small
\caption{Reader accuracy (\%) at a 32K context budget under overlap-union packing, averaged over seeds 5768, 78516, and 944601. LOONG uses Perfect Rate; NoLiMa is the custom diagnostic. Full and empty controls use all questions without a context budget. Confidence intervals and all budgets are reported in Appendix~\ref{app:reader-matrix}.}
\label{tab:reader-32k}
\begin{tabular}{lrrrr}
\toprule
Method & BABILong & LOONG & NoLiMa & LongMemEval \\
\midrule
Full context & 31.13 & 72.44 & 0.57 & 63.20 \\
Empty context & 15.00 & 0.00 & 0.00 & 10.27 \\
Head and tail & 21.40 & 78.67 & 0.00 & 25.27 \\
BM25 & 46.20 & 69.33 & 3.10 & 68.13 \\
Dense & 30.27 & 69.33 & 4.60 & 71.13 \\
Dense+Rerank & 32.20 & 75.56 & 5.29 & 70.60 \\
LLMLingua-2 & 12.07 & 31.56 & 0.57 & 43.00 \\
Provence & 48.13 & 35.11 & 27.43 & 66.60 \\
\bottomrule
\end{tabular}
\end{table}

\paragraph{Reader accuracy.}
Table~\ref{tab:reader-32k} reports three-run averages at 32K. On BABILong, BM25 reaches 46.20\% and Provence 48.13\%, versus 31.13\% for full context. Paired gains are 15.07 percentage points for BM25 (95\% CI: 10.87 to 19.60) and 17.00 for Provence (12.47 to 21.47). Provence minus BM25 is 1.93 points (\(-1.73\) to 5.73), which does not establish superiority. These contrasts are exploratory and unadjusted for multiple testing. Dense retrieval reaches 71.13\% on LongMemEval versus 63.20\% for full context. On LOONG, head and tail reaches 78.67\% Perfect Rate versus 72.44\% for full context. Perfect Rate requires a judge score of exactly 100. On the custom NoLiMa diagnostic, Provence reaches 27.43\% versus 0.57\% for full context. Appendix~\ref{app:reader-matrix} reports all budgets, judge settings, and 95\% paired-bootstrap intervals.

\section{Discussion and limitations}

Results are conditional on the evaluated cohorts, packing policy, and Gemini 2.5 Flash-Lite reader. Three seeds characterize this reader configuration. LOONG candidate lines are automatically identified; verbatim scoring can underestimate semantic retention, and automated answer metrics can err. NoLiMa is a custom diagnostic, and LongMemEval evaluates static selection. Shared token budgets differ from native model costs; recorded selection compute is not end-to-end latency. Appendix~\ref{app:reader-matrix} details scoring limitations, output caps, uncertainty, and validation.

\section{Conclusion}

\textsc{CoRA-Bench} evaluates evidence preservation and downstream answer quality on the same emitted contexts under shared context budgets. Evidence preservation favors lexical retrieval when surface cues are explicit and dense methods when evidence requires latent associations or conversational retrieval. The three-seed reader results yield different rankings: Provence has the highest 32K point estimates on BABILong and the custom NoLiMa diagnostic, while dense retrieval leads on LongMemEval. Differences between evidence recall and reader accuracy support reporting both metrics, particularly for text-modifying methods.

\clearpage
\begingroup
\small
\bibliographystyle{plainnat}
\bibliography{references}
\endgroup

\clearpage
\appendix

\section{Protocol details}
\label{app:protocol}

\paragraph{Packing and provenance.}
Evidence recall, token retention, and reader accuracy are computed on the same reader-visible contexts. BM25, dense retrieval, and dense reranking merge source-token spans to avoid duplicated overlap, preserve ranked-prefix selection and source order, and enforce the budget on the emitted text. Reranking uses $\min(n_{\mathrm{chunks}},\max(100,\lceil B/512\rceil+8))$ candidates. The fixed reranking candidate pool may leave part of the token budget unused after overlap removal. Selection records retain chunk identifiers, source offsets, token counts, and request identities. Evidence is mapped to source occurrences rather than a document-level flag. For methods that alter chunk text, provenance identifiers are ignored during scoring and the labeled text must remain verbatim.

\paragraph{LM filter.}
For LOONG, the LM-filter baseline uses 512-token chunks with 64-token overlap. Gemini 3 Flash Preview \citep{google2025gemini3flash} scores batches of 20 full chunks from 0 to 100 for relevance to the question, at temperature 0, seed 5768, and \texttt{thinkingLevel=MINIMAL}. This setting does not imply zero reasoning tokens. The \emph{Threshold $\geq50$, uncapped} arm keeps only chunks scoring at least 50, without filling a token budget. The \emph{ranked, capped} arms sort all chunks by score and pack a strict prefix up to 32K, 64K, or 128K, without applying the threshold. Both arms merge overlapping selected source spans and emit the retained text in source order, so shared regions appear only once. They reuse the same chunk relevance scores. Reader inference uses the common three-seed configuration with concurrency 4 and selection is scored once.

\paragraph{Selection compute.}
GPU-based selectors ran on Google Cloud \texttt{g2-standard-8} instances with one NVIDIA L4 (24\,GB) each, in \texttt{us-central1-\{a,b,c\}}, using PyTorch~2.9 and CUDA~12.9 \citep{nvidia2025cuda129}. Dense retrieval, reranking, LLMLingua-2, and Provence use GPU inference; BM25 and head-and-tail truncation are CPU-only. LM filtering and downstream inference use the Gemini API.

We measure selector compute-block wall time using \texttt{time.perf\_counter}. The measurements total approximately 48.5 L4 instance-hours: BABILong 17.5, NoLiMa 18.0, LongMemEval 9.4, and LOONG 3.6. Table~\ref{tab:compute} provides the per-method breakdown. Provence on NoLiMa required the most compute (11.5 L4 instance-hours), distributed across eight L4s.

For LM filtering, estimated API costs were \$9.80 for selection, \$3.98 for reader inference, and \$1.28 for judging. The recorded compute times exclude query encoding for dense retrieval, model loading, and downstream inference. They use the chunk-concatenation selection implementation and do not measure the complete overlap-union pipeline.

\begin{table}[htbp]
\centering
\caption{Historical per-method selection compute under chunk-concatenation packing, in L4 instance wall-clock hours (recorded via \texttt{time.perf\_counter} around each selector's main compute block; this is instance wall time, not measured GPU utilization). CPU-only baselines and API-based LM filtering are excluded. LM-filter API costs are reported separately in the text.}
\label{tab:compute}
\small
\begin{tabular}{lccccc}
\toprule
\textbf{Method} & \textbf{BABILong} & \textbf{LOONG} & \textbf{NoLiMa} & \textbf{LongMemEval} & \textbf{Total} \\
\midrule
Dense        & 2.5 & 0.3 & 1.2 & 1.1 & 5.1 \\
Dense+Rerank & 3.8 & 0.5 & 2.4 & 1.8 & 8.5 \\
LLMLingua-2  & 4.6 & 0.4 & 2.9 & 1.4 & 9.3 \\
Provence     & 6.6 & 2.4 & 11.5 & 5.1 & 25.6 \\
\midrule
Total        & 17.5 & 3.6 & 18.0 & 9.4 & 48.5 \\
\bottomrule
\end{tabular}
\end{table}

\paragraph{LM-filter API timing.}
The extension recorded 1,481 successful selector calls, 813 reader calls, and 248 distinct judge calls. Mean recorded request completion times were 1.57, 1.33, and 2.22 seconds, respectively. These per-request measurements include retries. They do not measure end-to-end latency. Provider-native usage under the recorded list prices determines the API cost estimates above.

\paragraph{Dataset licenses.}
BABILong code is Apache 2.0; its PG19 source is Apache 2.0 and bAbI is BSD. LOONG is Apache 2.0. The NoLiMa evaluation code and needle set use the Adobe Research License, which permits noncommercial research use; the five official haystacks retain the source-specific terms listed by NoLiMa. LongMemEval-cleaned is MIT. We used these assets for noncommercial research and do not redistribute upstream data.

\paragraph{Method and model licenses.}
The \texttt{rank-bm25} implementation is Apache 2.0. Qwen3-Embedding-0.6B and Qwen3-Reranker-0.6B are Apache 2.0. The LLMLingua software is MIT and the LLMLingua-2 MeetingBank checkpoint is Apache 2.0. The Provence checkpoint is CC BY-NC-ND 4.0 and was used only for noncommercial evaluation without modification or redistribution. \texttt{tiktoken} is MIT, PyTorch is BSD 3-Clause, and Transformers and Sentence Transformers are Apache 2.0. Gemini 3 Flash Preview was accessed through the official API under the Google Gemini API terms. Head and tail truncation is implemented locally and uses no external model asset.

\paragraph{Implementation and reproducibility.}
The implementation records dataset checksums, model identifiers, selection provenance, reader settings, scoring rules, and request identities. Configuration files and tests specify the packing, budget enforcement, and evaluation procedures. Source datasets and model weights remain subject to their owners' licenses.
\section{Reader evaluation}
\label{app:reader-matrix}

\paragraph{Repetitions and frozen cohorts.}
Each selected context and control is evaluated on the same questions at seeds \textbf{5768, 78516, and 944601}. BABILong data generation uses seed 42, independently of reader inference. The reader temperature is set to 0. Deterministic selectors are evaluated once per configuration, with reproducibility verified across three seeds on 30 prespecified questions per dataset. Selection equality is required across these checks. Identical reader requests may share a response only within a seed. Identical judge requests may share a judgment. Hence, the reported uncertainty excludes judge-repeat variability.

\begin{table}[ht]
\centering\small
\caption{Reader and evaluation settings. The same questions are evaluated at 3 seeds; dataset-native scoring rules are retained, with an adapted judge model for LOONG.}
\label{tab:reader-protocol}
\begin{tabularx}{\textwidth}{@{}lX@{}}
\toprule
Component & Configuration \\
\midrule
Reader & \texttt{gemini-2.5-flash-lite}; temperature 0; thinking budget 0; output cap 1,024; concurrency 8 \\
Reader repetitions & Same questions; seeds 5768, 78516, 944601 \\
BABILong metric & Pinned official task-label answer scorer and qa1-qa5 prompts \\
NoLiMa metric & Released case-sensitive contains scorer on the frozen custom diagnostic \\
LongMemEval judge & \texttt{gpt-4o-2024-08-06}; official question-type/abstention rubric; temperature 0; 10 output tokens \\
LOONG judge & \texttt{gpt-5.5-2026-04-23}; original rubric with an adapted judge model; reasoning disabled; 512 output tokens \\
LOONG metrics & Perfect Rate: score exactly 100; mean rubric score reported separately \\
Intervals & 2,000 paired bootstrap draws after averaging seeds within each question; BABILong task-stratified; NoLiMa needle-family-clustered \\
Latency pass & 30 matched questions per cell at reader seed 5768; selector controls use all 3 seeds; online concurrency 8 \\
\bottomrule
\end{tabularx}
\end{table}

\paragraph{Reference alignment and scoring.}
BABILong uses the released task-label first-sentence scorer. LOONG uses the published grading rubric and numeric parser with GPT-5.5 as an adapted judge in place of the reference GPT-4 evaluator. LongMemEval uses its question-type and abstention prompts with the reference GPT-4o snapshot. All 500 questions enter answer accuracy, including 30 abstentions. NoLiMa retains the released case-sensitive contains rule on our custom cohort.

\begin{figure}[p]
\centering
\includegraphics[width=\textwidth]{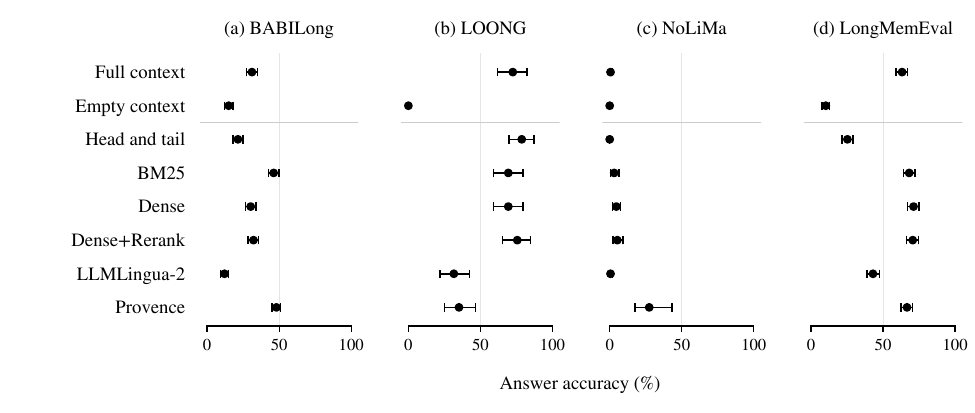}
\caption{Reader accuracy at 32K with Gemini 2.5 Flash-Lite, averaged over seeds 5768, 78516, and 944601. Points show mean accuracy. Horizontal bars are 95\% paired-bootstrap intervals after averaging reader runs within question. BABILong is task-stratified and NoLiMa is needle-family-clustered. Full/empty controls have no budget cap. LOONG reports Perfect Rate; other panels report their dataset-specific answer metrics. Evidence and token-retention measurements in Figure~\ref{fig:pareto-frontiers} use these same emitted contexts.}
\label{fig:reader-accuracy}
\end{figure}

\paragraph{Intervals and seed variation.}
For each question we first average correctness over the 3 reader runs, then resample paired questions using 2,000 bootstrap draws and report the 2.5th and 97.5th percentiles. BABILong draws are stratified by task. NoLiMa draws keep related variants, depths, and haystacks together within a needle family (ten clusters). The same draws support paired method differences, reported as exploratory and unadjusted for multiple comparisons. Reader seeds are averaged within question, not treated as independent observations. Tables additionally report the minimum and maximum of the 3 seed accuracies; those ranges are not confidence intervals. Figure~\ref{fig:reader-accuracy} visualizes the 32K comparison. Configurations without complete 3-seed scoring are excluded from inferential comparisons.

\paragraph{Coverage and validation.}
All 68 configurations in the completed six-method matrix and controls have complete 3-seed scoring: 91,518 logical answer outcomes and 73,272 distinct saved reader receipts. The separate LM-filter extension adds four LOONG configurations and 813 fully judged reader outcomes from new inference on all 75 sources. All 28,709 chunks have valid scores; all selections, evidence scores, reader requests, and native judge parses were reconstructed from saved artifacts. We validated source and request identities, emitted contexts, budget compliance, and scoring outputs across the primary matrix, with additional sampled receipt reconstruction. The tables report scoring coverage, output-cap counts, and complementary LOONG mean rubric scores.

\paragraph{Scope and limitations.}
LOONG uses Financial Spotlight level 1 and budget-specific cohorts. NoLiMa is a custom repeated-measures diagnostic with ten needle-family clusters, rather than the official aggregate or its base-length normalization. LongMemEval evaluates static context selection, rather than online memory management. Automatically identified LOONG candidate lines may be ambiguous, and verbatim scoring can underestimate semantic retention by text-modifying methods. String matching may accept unsupported answer mentions, automated judges may err, and source/reference unit ambiguities can affect financial scores. We retain each dataset's stated scoring rules.

Reader evaluations use Gemini 2.5 Flash-Lite with thinking disabled and a 1,024-token output cap. We report output-cap hits and quantify variation across three seeds for this reader configuration. Confidence intervals reflect sampling variability within the evaluated cohorts, with the stratification and clustering described above. The evidence tables report fixed-cohort point estimates. The selection-compute totals in Appendix~\ref{app:protocol} exclude query encoding and cannot be interpreted as end-to-end serving latency.

Future evaluations could extend this protocol to online memory systems that update stored information across interactions \citep{park2023generative,zhong2024memorybank,packer2023memgpt}.

\paragraph{Token accounting and tables.}
Budgets count the selected context using \texttt{cl100k\_base}, excluding task prompt overhead. Provider-native input/output usage determines reader API cost, and the native reader window is checked separately. Chunk length in the shared tokenizer is not assumed to equal native model length; the completed audit covers 671,634 source chunks for Provence preprocessing and embedding/reranker input-length limits. LLMLingua-2's internal segmentation is outside this audit. Provence windows cover long passages, while its ranking score uses the first native block. We retain raw materialized table text and apply no table-specific preservation, routing, or whitespace repair.

\paragraph{Hardware and timing scope.}
The reader evaluation uses an NVIDIA L4 (24\,GB) per \texttt{g2-standard-8} GPU worker and \texttt{e2-standard-4} CPU execution. A complete standalone selection measurement includes chunking, query encoding, inference, and packing, with CUDA synchronization and fresh data caches. Incremental timings that reuse work across budgets are not standalone latency estimates. Provider-native input and output usage determines API cost.

\begin{table}[p]
\centering
\small
\setlength{\tabcolsep}{3pt}
\caption{BABILong: reader evaluation under overlap-union packing. Accuracy is a percentage; brackets are 95\% paired-bootstrap intervals after averaging three reader runs within each question (BABILong: task-stratified; NoLiMa: needle-family-clustered). The seed range is descriptive, not a confidence interval. Coverage counts questions scored at all three seeds; cap hits count outputs reaching the output-token limit.}
\label{tab:reader-babilong}
\begin{tabular}{llrrrr}
\toprule
Method & Budget & Coverage & Accuracy [95\% CI] & Seed min-max & Cap hits \\
\midrule
Full context & n/a & 500/500 & 31.13 [27.33, 35.07] & 30.80 to 31.60 & 0 \\
Empty context & n/a & 500/500 & 15.00 [12.00, 18.00] & 14.20 to 15.40 & 0 \\
Head and tail & 32K & 500/500 & 21.40 [18.00, 24.80] & 21.40 to 21.40 & 0 \\
Head and tail & 64K & 500/500 & 30.13 [26.33, 33.87] & 30.00 to 30.20 & 0 \\
Head and tail & 128K & 500/500 & 32.47 [28.53, 36.40] & 32.40 to 32.60 & 0 \\
BM25 & 32K & 500/500 & 46.20 [42.60, 49.80] & 46.20 to 46.20 & 0 \\
BM25 & 64K & 500/500 & 45.47 [41.67, 49.00] & 45.00 to 46.00 & 0 \\
BM25 & 128K & 500/500 & 39.47 [35.60, 43.47] & 39.40 to 39.60 & 0 \\
Dense & 32K & 500/500 & 30.27 [26.60, 34.07] & 30.20 to 30.40 & 0 \\
Dense & 64K & 500/500 & 32.53 [28.73, 36.33] & 32.00 to 33.00 & 0 \\
Dense & 128K & 500/500 & 32.73 [28.87, 36.47] & 32.00 to 33.20 & 0 \\
Dense+Rerank & 32K & 500/500 & 32.20 [28.40, 35.80] & 32.20 to 32.20 & 0 \\
Dense+Rerank & 64K & 500/500 & 32.20 [28.33, 35.93] & 31.80 to 32.60 & 0 \\
Dense+Rerank & 128K & 500/500 & 33.47 [29.47, 37.33] & 33.00 to 34.00 & 0 \\
LLMLingua-2 & 32K & 500/500 & 12.07 [9.40, 14.87] & 12.00 to 12.20 & 0 \\
LLMLingua-2 & 64K & 500/500 & 14.27 [11.27, 17.20] & 14.20 to 14.40 & 0 \\
LLMLingua-2 & 128K & 500/500 & 16.60 [13.53, 19.93] & 16.40 to 16.80 & 0 \\
Provence & 32K & 500/500 & 48.13 [45.20, 50.93] & 48.00 to 48.20 & 0 \\
Provence & 64K & 500/500 & 48.13 [45.20, 50.93] & 48.00 to 48.20 & 0 \\
Provence & 128K & 500/500 & 48.13 [45.20, 50.93] & 48.00 to 48.20 & 0 \\
\bottomrule
\end{tabular}
\end{table}
\begin{table}[p]
\centering
\small
\setlength{\tabcolsep}{3pt}
\caption{LOONG: reader evaluation under overlap-union packing. The LM-filter threshold arm keeps scores $\geq50$ without a cap; its ranked arms use no score threshold. Accuracy is LOONG Perfect Rate with the adapted GPT-5.5 judge; eligible cohorts are 75/70/51 at 32K/64K/128K; brackets are 95\% paired-bootstrap intervals after averaging three reader runs within each question (BABILong: task-stratified; NoLiMa: needle-family-clustered). The seed range is descriptive, not a confidence interval. Coverage counts questions scored at all three seeds; cap hits count outputs reaching the output-token limit.}
\label{tab:reader-loong}
\resizebox{\textwidth}{!}{%
\begin{tabular}{llrrrr}
\toprule
Method & Budget / policy & Coverage & Accuracy [95\% CI] & Seed range & Cap hits \\
\midrule
Full context & n/a & 75/75 & 72.44 [61.78, 82.22] & 72.00 to 73.33 & 0 \\
Empty context & n/a & 75/75 & 0.00 [0.00, 0.00] & 0.00 to 0.00 & 0 \\
Head and tail & 32K & 75/75 & 78.67 [69.78, 87.12] & 77.33 to 80.00 & 0 \\
Head and tail & 64K & 70/70 & 80.00 [70.48, 88.57] & 78.57 to 81.43 & 0 \\
Head and tail & 128K & 51/51 & 67.32 [54.90, 79.08] & 64.71 to 70.59 & 0 \\
BM25 & 32K & 75/75 & 69.33 [59.11, 79.56] & 68.00 to 70.67 & 0 \\
BM25 & 64K & 70/70 & 73.81 [62.86, 82.86] & 72.86 to 74.29 & 0 \\
BM25 & 128K & 51/51 & 60.13 [46.41, 73.86] & 58.82 to 60.78 & 0 \\
Dense & 32K & 75/75 & 69.33 [59.10, 79.56] & 69.33 to 69.33 & 0 \\
Dense & 64K & 70/70 & 71.90 [61.43, 81.90] & 71.43 to 72.86 & 0 \\
Dense & 128K & 51/51 & 62.09 [49.02, 75.16] & 58.82 to 64.71 & 0 \\
Dense+Rerank & 32K & 75/75 & 75.56 [65.33, 84.89] & 74.67 to 76.00 & 0 \\
Dense+Rerank & 64K & 70/70 & 70.00 [58.57, 81.43] & 70.00 to 70.00 & 0 \\
Dense+Rerank & 128K & 51/51 & 60.78 [47.71, 74.51] & 58.82 to 62.75 & 0 \\
LLMLingua-2 & 32K & 75/75 & 31.56 [21.78, 42.22] & 30.67 to 32.00 & 1 \\
LLMLingua-2 & 64K & 70/70 & 42.86 [31.43, 54.29] & 42.86 to 42.86 & 0 \\
LLMLingua-2 & 128K & 51/51 & 58.17 [45.10, 71.90] & 56.86 to 58.82 & 0 \\
Provence & 32K & 75/75 & 35.11 [24.89, 46.67] & 34.67 to 36.00 & 0 \\
Provence & 64K & 70/70 & 36.19 [25.71, 47.62] & 35.71 to 37.14 & 0 \\
Provence & 128K & 51/51 & 35.29 [21.57, 49.02] & 35.29 to 35.29 & 0 \\
LM filter & \shortstack{Threshold $\geq50$,\\uncapped} & 75/75 & 54.67 [44.00, 65.33] & 54.67 to 54.67 & 0 \\
LM filter (ranked) & 32K & 75/75 & 74.67 [65.33, 84.00] & 74.67 to 74.67 & 0 \\
LM filter (ranked) & 64K & 70/70 & 75.71 [65.71, 85.71] & 75.71 to 75.71 & 0 \\
LM filter (ranked) & 128K & 51/51 & 63.40 [50.98, 76.47] & 62.75 to 64.71 & 0 \\
\bottomrule
\end{tabular}
}
\end{table}

\begin{table}[p]
\centering
\small
\setlength{\tabcolsep}{3pt}
\caption{NoLiMa custom 870-observation diagnostic: reader evaluation under overlap-union packing. Accuracy is a percentage; brackets are 95\% paired-bootstrap intervals after averaging three reader runs within each question (BABILong: task-stratified; NoLiMa: needle-family-clustered). The seed range is descriptive, not a confidence interval. Coverage counts questions scored at all three seeds; cap hits count outputs reaching the output-token limit.}
\label{tab:reader-nolima}
\begin{tabular}{llrrrr}
\toprule
Method & Budget & Coverage & Accuracy [95\% CI] & Seed min-max & Cap hits \\
\midrule
Full context & n/a & 870/870 & 0.57 [0.00, 1.48] & 0.57 to 0.57 & 0 \\
Empty context & n/a & 870/870 & 0.00 [0.00, 0.00] & 0.00 to 0.00 & 0 \\
Head and tail & 32K & 870/870 & 0.00 [0.00, 0.00] & 0.00 to 0.00 & 0 \\
Head and tail & 64K & 870/870 & 0.27 [0.00, 0.88] & 0.23 to 0.34 & 0 \\
BM25 & 32K & 870/870 & 3.10 [0.65, 6.42] & 3.10 to 3.10 & 0 \\
BM25 & 64K & 870/870 & 1.69 [0.17, 4.11] & 1.61 to 1.72 & 0 \\
Dense & 32K & 870/870 & 4.60 [1.81, 7.45] & 4.60 to 4.60 & 0 \\
Dense & 64K & 870/870 & 2.64 [0.66, 5.57] & 2.64 to 2.64 & 0 \\
Dense+Rerank & 32K & 870/870 & 5.29 [2.29, 9.14] & 5.17 to 5.40 & 0 \\
Dense+Rerank & 64K & 870/870 & 2.84 [0.88, 5.61] & 2.76 to 2.87 & 0 \\
LLMLingua-2 & 32K & 870/870 & 0.57 [0.00, 2.08] & 0.57 to 0.57 & 0 \\
LLMLingua-2 & 64K & 870/870 & 0.34 [0.00, 0.87] & 0.34 to 0.34 & 0 \\
Provence & 32K & 870/870 & 27.43 [17.65, 43.08] & 27.36 to 27.47 & 0 \\
Provence & 64K & 870/870 & 27.43 [17.65, 43.08] & 27.36 to 27.47 & 0 \\
\bottomrule
\end{tabular}
\end{table}

\begin{table}[p]
\centering
\small
\setlength{\tabcolsep}{3pt}
\caption{LongMemEval: reader evaluation under overlap-union packing. Accuracy includes all 500 questions, including 30 abstentions, and is a percentage; brackets are 95\% paired-bootstrap intervals after averaging three reader runs within each question (BABILong: task-stratified; NoLiMa: needle-family-clustered). The seed range is descriptive, not a confidence interval. Coverage counts questions scored at all three seeds; cap hits count outputs reaching the output-token limit.}
\label{tab:reader-longmemeval}
\begin{tabular}{llrrrr}
\toprule
Method & Budget & Coverage & Accuracy [95\% CI] & Seed min-max & Cap hits \\
\midrule
Full context & n/a & 500/500 & 63.20 [59.00, 67.13] & 62.80 to 63.80 & 156 \\
Empty context & n/a & 500/500 & 10.27 [7.53, 13.00] & 10.20 to 10.40 & 3 \\
Head and tail & 32K & 500/500 & 25.27 [21.53, 29.07] & 25.00 to 25.40 & 97 \\
Head and tail & 64K & 500/500 & 44.13 [39.93, 48.20] & 44.00 to 44.20 & 121 \\
BM25 & 32K & 500/500 & 68.13 [64.00, 72.07] & 67.00 to 68.80 & 115 \\
BM25 & 64K & 500/500 & 67.87 [63.93, 71.93] & 67.60 to 68.20 & 118 \\
Dense & 32K & 500/500 & 71.13 [67.07, 75.00] & 71.00 to 71.40 & 128 \\
Dense & 64K & 500/500 & 68.40 [64.33, 72.27] & 67.20 to 69.60 & 118 \\
Dense+Rerank & 32K & 500/500 & 70.60 [66.33, 74.60] & 70.20 to 71.00 & 110 \\
Dense+Rerank & 64K & 500/500 & 67.27 [63.33, 71.20] & 67.00 to 67.60 & 143 \\
LLMLingua-2 & 32K & 500/500 & 43.00 [39.00, 47.40] & 43.00 to 43.00 & 429 \\
LLMLingua-2 & 64K & 500/500 & 56.33 [52.07, 60.67] & 56.00 to 57.00 & 259 \\
Provence & 32K & 500/500 & 66.60 [62.60, 70.60] & 66.60 to 66.60 & 51 \\
Provence & 64K & 500/500 & 66.60 [62.60, 70.60] & 66.60 to 66.60 & 51 \\
\bottomrule
\end{tabular}
\end{table}
\begin{table}[p]
\centering
\small
\caption{LOONG complementary mean rubric scores (0 to 100 scale as returned by the native parser) with the adapted GPT-5.5 judge at each budget. Full and empty controls use all 75 questions. The LM-filter ranked arms apply no threshold; Threshold $\geq50$, uncapped has mean score 81.15 on all 75 questions. Scores outside the requested 1 to 100 rubric range are preserved and flagged, not clipped.}
\label{tab:reader-loong-scores}
\begin{tabular}{lrrr}
\toprule
Method & 32K & 64K & 128K \\
\midrule
Full context & 88.75 & 88.75 & 88.75 \\
Empty context & 6.60 & 6.60 & 6.60 \\
Head and tail & 91.21 & 90.39 & 87.19 \\
BM25 & 81.11 & 85.78 & 79.48 \\
Dense & 84.39 & 88.72 & 81.73 \\
Dense+Rerank & 89.98 & 88.67 & 82.45 \\
LLMLingua-2 & 44.52 & 61.06 & 78.81 \\
Provence & 57.80 & 57.71 & 55.69 \\
LM filter (ranked) & 88.72 & 88.66 & 84.50 \\
\bottomrule
\end{tabular}
\end{table}

\clearpage

\section{Evidence preservation and compression}
\label{app:results}

Tables~\ref{tab:babilong-appendix}-\ref{tab:lme-appendix} report evidence recall and compression ratio on the same emitted contexts used in Appendix~\ref{app:reader-matrix}. Both metrics are macro question-level means, with evidence evaluated only where annotations are available. Lower compression ratios indicate fewer transmitted tokens. Text-modifying methods use verbatim evidence survival; extractive methods use source occurrence or span provenance.

\begin{table}[ht]
\centering
\caption{BABILong macro supporting-fact recall and macro compression ratio. Each cell uses the same 500 examples, all longer than 128K tokens.}
\label{tab:babilong-appendix}
\small
\setlength{\tabcolsep}{4pt}
\begin{tabular}{lcccccc}
\toprule
& \multicolumn{3}{c}{\textbf{Evidence recall}} & \multicolumn{3}{c}{\textbf{Compression ratio}} \\
\cmidrule(lr){2-4}\cmidrule(lr){5-7}
\textbf{Method} & 32K & 64K & 128K & 32K & 64K & 128K \\
\midrule
Head and tail & 0.238 & 0.446 & 0.659 & 0.130 & 0.260 & 0.519 \\
BM25 & \textbf{0.838} & \textbf{0.868} & \textbf{0.921} & 0.129 & 0.259 & 0.518 \\
Dense & 0.329 & 0.481 & 0.727 & 0.129 & 0.259 & 0.519 \\
Dense+Rerank & 0.395 & 0.483 & 0.714 & 0.129 & 0.258 & 0.491 \\
LLMLingua-2 & 0.000 & 0.000 & 0.003 & 0.105 & 0.216 & 0.453 \\
Provence & 0.750 & 0.750 & 0.750 & 0.014 & 0.014 & 0.014 \\
\bottomrule
\end{tabular}
\end{table}

\begin{table}[ht]
\centering
\caption{LOONG macro automatically matched candidate-line survival and macro compression ratio. Evidence denominators are 74/69/50; reader and token denominators are 75/70/51 at 32K/64K/128K. LM filter (ranked) applies no score threshold. Threshold $\geq50$, uncapped is reported separately: survival 0.932 ($n=74$) and compression ratio 0.037 ($n=75$).}
\label{tab:loong-appendix}
\small
\setlength{\tabcolsep}{4pt}
\begin{tabular}{lcccccc}
\toprule
& \multicolumn{3}{c}{\textbf{Evidence recall}} & \multicolumn{3}{c}{\textbf{Compression ratio}} \\
\cmidrule(lr){2-4}\cmidrule(lr){5-7}
\textbf{Method} & 32K & 64K & 128K & 32K & 64K & 128K \\
\midrule
Head and tail & 0.973 & 0.986 & \textbf{1.000} & 0.241 & 0.427 & 0.653 \\
BM25 & 0.932 & 0.942 & 0.980 & 0.239 & 0.426 & 0.652 \\
Dense & 0.946 & 0.986 & 0.980 & 0.240 & 0.425 & 0.652 \\
Dense+Rerank & 0.973 & 0.986 & 0.980 & 0.240 & 0.417 & 0.609 \\
LLMLingua-2 & 0.122 & 0.188 & 0.220 & 0.226 & 0.389 & 0.585 \\
Provence & 0.676 & 0.667 & 0.640 & 0.017 & 0.017 & 0.017 \\
LM filter (ranked) & \textbf{1.000} & \textbf{1.000} & \textbf{1.000} & 0.240 & 0.425 & 0.652 \\
\bottomrule
\end{tabular}
\end{table}

\begin{table}[ht]
\centering
\caption{Custom NoLiMa normalized verbatim needle survival and macro compression ratio over 870 repeated observations.}
\label{tab:nolima-appendix}
\small
\begin{tabular}{lcccc}
\toprule
& \multicolumn{2}{c}{\textbf{Evidence recall}} & \multicolumn{2}{c}{\textbf{Compression ratio}} \\
\cmidrule(lr){2-3}\cmidrule(lr){4-5}
\textbf{Method} & 32K & 64K & 32K & 64K \\
\midrule
Head and tail & 0.000 & 0.333 & 0.249 & 0.498 \\
BM25 & 0.264 & 0.467 & 0.247 & 0.496 \\
Dense & 0.641 & \textbf{0.908} & 0.247 & 0.496 \\
Dense+Rerank & \textbf{0.683} & \textbf{0.908} & 0.247 & 0.495 \\
LLMLingua-2 & 0.000 & 0.000 & 0.210 & 0.450 \\
Provence & 0.332 & 0.332 & 0.005 & 0.005 \\
\bottomrule
\end{tabular}
\end{table}

\begin{table}[ht]
\centering
\caption{LongMemEval macro answer-bearing turn recall over 470 answerable examples and macro compression ratio over all 500 examples, including 30 abstentions. All methods have complete coverage.}
\label{tab:lme-appendix}
\small
\begin{tabular}{lcccc}
\toprule
& \multicolumn{2}{c}{\textbf{Evidence recall}} & \multicolumn{2}{c}{\textbf{Compression ratio}} \\
\cmidrule(lr){2-3}\cmidrule(lr){4-5}
\textbf{Method} & 32K & 64K & 32K & 64K \\
\midrule
Head and tail & 0.301 & 0.645 & 0.305 & 0.609 \\
BM25 & 0.940 & 0.976 & 0.302 & 0.607 \\
Dense & 0.967 & \textbf{0.993} & 0.302 & 0.607 \\
Dense+Rerank & \textbf{0.979} & 0.991 & 0.302 & 0.587 \\
LLMLingua-2 & 0.000 & 0.000 & 0.286 & 0.600 \\
Provence & 0.036 & 0.036 & 0.049 & 0.049 \\
\bottomrule
\end{tabular}
\end{table}

\subsection{Analysis of evidence preservation}

\paragraph{Method rankings depend on the retrieval regime.}
On BABILong, BM25 has the highest evidence recall at every budget, reaching 0.838, 0.868, and 0.921. Dense retrieval improves from 0.329 to 0.727 as the budget grows, while reranking raises the 32K result to 0.395. On the custom NoLiMa diagnostic, where literal overlap is intentionally limited, the ordering changes. Dense and Dense+Rerank reach 0.641 and 0.683 at 32K, compared with 0.264 for BM25. At 64K, dense and Dense+Rerank both reach 0.908. These results support a conditional conclusion: lexical retrieval is effective when benchmark templates expose entity overlap, while semantic retrieval is more robust to a lexical gap.

\paragraph{Retrieval transfers to static dialogue selection.}
On LongMemEval, Dense+Rerank retains 0.979 and 0.991 of answer-bearing turns at 32K and 64K. Dense retrieval reaches 0.967 and 0.993, and BM25 reaches 0.940 and 0.976. Head and tail truncation is substantially lower at 0.301 and 0.645. At 32K, reranking therefore preserves nearly all annotated conversational evidence while transmitting about 0.302 of the original tokens. Dense retrieval has the highest evidence-recall point estimate at 64K.

\paragraph{Compression exposes distinct notions of preservation.}
On BABILong, Provence retains 0.750 of supporting facts while transmitting approximately 0.014 of the original tokens. Its complete-turn recall on LongMemEval is 0.036 under verbatim scoring, while LLMLingua-2 rarely preserves the complete labeled strings on BABILong, NoLiMa, or LongMemEval. Because both methods transform text, this result identifies a measurable gap between exact evidence survival and possible semantic sufficiency. Reader accuracy provides a complementary measure of semantic utility, on the same emitted contexts, subject to the reader configuration.

\paragraph{LOONG and the LM filter diagnostic.}
On LOONG, head and tail truncation retains 0.973, 0.986, and 1.000 of automatically matched candidate lines. Dense+Rerank reaches 0.973, 0.986, and 0.980, while BM25 reaches 0.932, 0.942, and 0.980. LLMLingua-2 ranges from 0.122 to 0.220, and Provence ranges from 0.640 to 0.676 while using about 0.017 of source tokens. The LM-filter \emph{Threshold $\geq50$, uncapped} arm achieves 0.932 candidate-line survival, retains 0.037 of source tokens, and reaches 54.67\% reader Perfect Rate (mean rubric score 81.15/100). The ranked, capped arms achieve 1.000 candidate-line survival at all three budgets, retain 0.240, 0.425, and 0.652 of source tokens, and reach 74.67\%, 75.71\%, and 63.40\% Perfect Rate. Their larger outputs reflect ranking without the threshold, rather than a cap applied to the threshold-selected context. Accuracy and uncertainty appear in Table~\ref{tab:reader-loong}; cross-budget differences also reflect the eligible cohorts.

\subsection{Recall-compression tradeoffs}

\begin{figure}[ht]
\centering
\includegraphics[width=\textwidth]{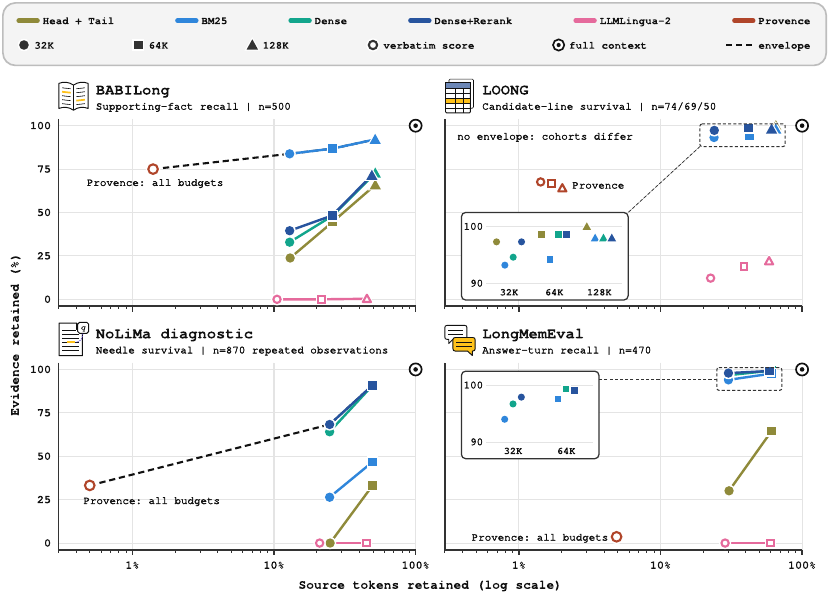}
\caption{Evidence recall versus retained source tokens across four datasets. Both axes show percentages, with a logarithmic token axis. Marker shapes indicate token budgets with hollow markers denoting verbatim evidence scoring and ringed markers denoting full context. Dashed lines show nondominated envelopes for fixed-cohort datasets. LOONG uses different cohorts across budgets (74/69/50 evidence-evaluable examples). Insets enlarge clustered results, with horizontally offset markers for clarity. Provence is shown once where results coincide across budgets. LM-filter results are excluded. All values are point estimates without confidence intervals.}
\label{fig:pareto-frontiers}
\end{figure}

Figure~\ref{fig:pareto-frontiers} shows evidence recall against retained context for the six non-LLM selectors. LOONG LM-filter results are reported in the accompanying analysis.

\end{document}